\documentclass[runningheads]{llncs}
\usepackage{graphicx}
\usepackage{pdfpages}
\usepackage{amsmath}
\usepackage{tikz}
\usepackage{multirow}%new
\usepackage{setspace}%new
\usepackage{cases}%new
\usepackage{arydshln}
\usepackage{algorithm}
\usepackage{algorithmic}
\usepackage{booktabs}

\DeclareMathOperator*{\argmax}{arg\,max}
\newcommand{\tabincell}[2]{\begin{tabular}{@{}#1@{}}#2\end{tabular}} 

\begin{document}
\title{A Corpus-free State2Seq User Simulator for Task-oriented Dialogue}

\author{Yutai Hou$^{\dag}$, Meng Fang$^{\ddag}$, Wanxiang Che\thanks{\ ~Email correspondence.}$^{\dag}$, Ting Liu$^{\dag}$ \\
	$^{\dag}$Research Center for Social Computing and Information Retrieval \\
	Harbin Institute of Technology, China \\
	{\tt \{ythou, car, tliu\}@ir.hit.edu.cn} \\
	$^{\ddag}$ Tencent Robotics X, {\tt mfang@tencent.com}
}

\institute{}

\maketitle              % typeset the header of the contribution
\begin{abstract}
Recent reinforcement learning algorithms for task-oriented dialogue system absorbs a lot of interest. 
However,
an unavoidable obstacle for training such algorithms is that annotated dialogue corpora are often unavailable.
One of the popular approaches addressing this is to train a dialogue agent with a user simulator. 
Traditional user simulators are built upon a set of dialogue rules and therefore lack response diversity. 
This severely limits the simulated cases for agent training. 
Later data-driven user models work better in diversity but suffer from data scarcity problem. 
To remedy this, we design a new corpus-free framework that taking advantage of their benefits. 
The framework builds a user simulator by first generating diverse dialogue data from templates and 
then build a new State2Seq user simulator on the data. 
To enhance the performance, 
we propose the State2Seq user simulator model to efficiently leverage dialogue state and history. 
%Based on the framework, it is very efficient to train dialogue agents with low cost.
Experiment results on an open dataset show that 
our user simulator helps agents achieve an improvement of  $6.36\%$ on success rate. 
State2Seq model outperforms the seq2seq baseline for $1.9$ F-score. 

%\keywords{First keyword  \and Second keyword \and Another keyword.}
\end{abstract}
\section{Introduction}
%\atmasay{Overall logic
%	1 Policy could be got just with user simulator 
%	2 But when there is no data, you can't get the data-driven user simulator.
%	3 So we draw on the idea of self-play and propose a new method of training user simulator.
%	4 At the same time, based on the work of our predecessors, we proposed a new user simulator State2Seq, and achieved significant improvements.
%}

%Idea logic:
%Intro to task oriented dialogue
%User simulator is important
%Previous environment
%Our method overview
%Data generation process detail
%User simulator detail
%Experiment result

Task-oriented dialogue systems assist users to achieve specific goals 
such as finding restaurants or booking flights \cite{young2013pomdp}. 
To learn such a system is very challenging. 
Recently, reinforcement learning (RL) methods have been introduced 
due to their advantages in sequential decision. 
%to build task-oriented dialogue agents
\cite{young2013pomdp,DBLP:conf/acl/RoyPT00,williams2007partially,gavsic2014gaussian}. 
%RL methods allow to optimize dialogue policy without any expert-generated examples. 
An RL based dialogue agent can learn from dialogue data or reward signals by interacting with real users. 
Unfortunately, interacting with real users is costly and 
there is often no enough data or even no data for new domains. 
To overcome these obstacles, 
building user simulators is studied for training RL dialogue algorithms \cite{SCHATZMANN2006,li2016user}. 

\begin{figure}[t]
	\small
	\centering\includegraphics[width=0.8\columnwidth]{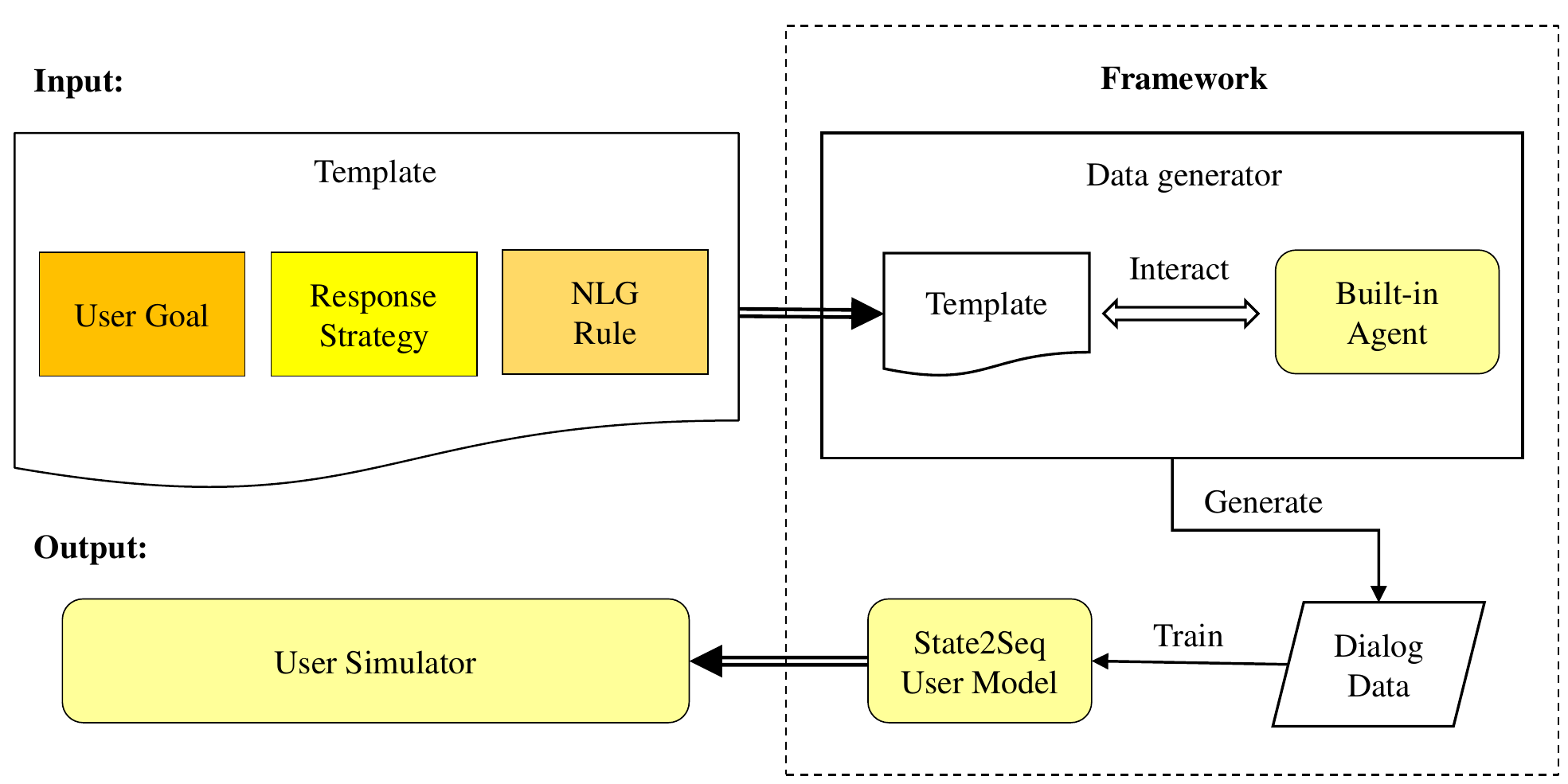}
	\caption{The proposed corpus-free framework for building user simulators.}\label{fig:framework}
%	\vspace*{-5mm}
\end{figure}

User simulators can be divided into two categories: traditional and data-driven user simulator. 
Traditional user simulators are agenda-based or rule-based \cite{Li2017,schatzmann2007agenda}. 
A rule-based user simulator can be built without data, 
but needs domain-specific knowledge and hard to generalize to new contexts. 
Besides, the rule-based model lacks response diversity, 
which largely limits the effectiveness of RL training. 
Latter data-driven user simulators ease the problem of diversity and depend less on expert knowledge. 
They imitate user behaviors from datasets with statistical models, 
such as Bayesian models \cite{pietquin2006probabilistic,georgila2005learning}, 
hidden Markov models \cite{cuayahuitl2005human} and seq2seq models \cite{DBLP:conf/interspeech/AsriHS16}. 
Statistical user simulators are inherently diverse and often require a large amount of expert-labeled data for training. 
However, 
%collecting and annotating human-machine or human-human conversations for task-oriented dialogues is expensive and time-consuming and requires extensive domain knowledge. 
%So there is often no enough data or even no data, especially when building dialogue systems for new domains. 
they can not cope with limited data situations. 

In this paper, 
to overcome the data scarcity problem and build a user simulator with sufficient diversity, 
we propose a new corpus-free framework for building user simulators. 
It combines the ideas of rule-based and data-driven user simulators. 
%Our framework leverages the bottleneck of data limitation by introducing predefined templates
%and then learns a diversity user simulator with a new State2Seq attention mechanism.
As shown in Figure 1, the framework generates dialogue data from templates and train data-driven user model on it.
The template consists of user goals, response strategy and natural language generation rules. 
An example of the template is shown in Table \ref{tbl:template}. 
%It is used for building a user simulator from scratch. 
%In order to avoid lacking diversity and generalization, 
In addition to templates, data generation uses an RL-based built-in agent to improve data diversity and explore more dialogue cases. 
The statistical nature of data-driven user simulator provides more diversity than rule-based ones. 
Diverse responses allow covering more situation for policy training. 

To enhance the user simulator's quality, 
we propose a novel attention based State2Seq user simulator to leverage the dialogue state and history better. 
The model first learns representations for dialogue context items. 
Dialogue context contains structured data of dialogue states, user goals and agent response. 
%We define dialogue states in the feature style of \cite{DBLP:conf/interspeech/AsriHS16}. 
%Then the model generates user response by paying attention to 
Then for each dialogue turn, the model predicts user actions sequentially with the attention on context. 
Attention helps to pick action more accurately. 
For example,  suppose agent response is (request:movie, inform:date=today), 
and user goal is (request=[ticket],  constraint=[date:tomorrow, movie:Deadpool]). 
The user model can easily output actions (inform:movie, deny:date) by attending to the agent request and the states of constraint inconsistency respectively. 
%the representations of $<$user goal, dialogue state, agent response$>$. 

Experiments are conducted on the movie booking dataset \cite{Li2017} and an in-house restaurant domain dataset. 
%We also construct a dataset of restaurant domainFor generalization verification.. 
We evaluate both the user simulator model itself and the policy trained by it. 
%the performance of agent trained by it, and the validation of the user simulator. 
%Results show the proposed method outperforms baselines.
On movie booking dataset, 
our policy achieves an improvement of  $6.36$ points on the success rate over the strongest baseline. 
%our policy outperform the strongest baseline for $6.36$ points on success rate
And proposed State2Seq model outperforms the seq2seq baseline for $1.9$ F-score on response accuracy. 

%There are 3 main contributions of the paper: 
This paper has 3 main contributions: 
1. To solve data scarcity, we design a new corpus-free framework for building a user simulator with response diversity.
2. We introduce the attention mechanism to task-oriented user simulator and 
propose a State2Seq model to get better user behavior modeling. 
3. Experiments show that 
response diversity and attention on dialogue context improve user model and agent policy. 

Our code is available at: 
\url{https://github.com/AtmaHou/UserSimulator}

%\section{Method}
\section{Proposed Framework}

We focus on developing a user simulator, 
which is diverse to cover enough dialogue situations.
To solve the data scarcity problem, our framework builds a user simulator with only templates and no dialogue data. 
There are two main components in the framework:
a template based data generator and a neural user simulator. 
The framework 
(1) first generate data from the templates using a built-in agent, 
(2) then train a data-driven user simulator on it. 

\begin{table}[t]
	\centering
	\renewcommand\arraystretch{1.5}
	%	\tiny
	\scriptsize 
	%	\footnotesize
	%	\small
	\caption{Template example for movie booking domain.
		$G$ is user goal, $V$ is response strategy and $N$ is NLG rules. \texttt{req} is request.}
	\begin{tabular}{crp{5cm}l}
		\toprule
		\textbf{Template Name} & \multicolumn{3}{c}{\textbf{Template Content}} \\
		\hline
		\multirow{2}{*}{G} & \multirow{2}{*}{$ g_0$ =  $\bigg [$} & $C$ =[ \texttt{movie} = Godfather, \texttt{time} = 5 pm ],  &   \multirow{2}{*}{$\bigg ]$, $g_1, g_2, ... $} \\
		&																	& $R$ = [\,\texttt{ticket}, \texttt{theater}] & \\
%		\hdashline
		\multirow{2}{*}{V} & \multirow{2}{*}{$r_0$ =  $\bigg [$} & \textbf{if} $A_{t-1}$ = \{\texttt{req:time}\} and \texttt{time} in $g.C$ & \multirow{2}{*}{$\bigg ]$, $r_1, r_2, ... $} \\
		&																	& \textbf{then} $A_t$ = {\texttt{inform:time} = $g.C.\texttt{time}$}  &  \\
%		\hdashline
		\multirow{2}{*}{N} & \multirow{2}{*}{$l_0$ =  $\bigg [$} & \textbf{if} $A$ = \{\texttt{inform:time}=g.C.\texttt{time}\}  & \multirow{2}{*}{$\bigg ]$, $l_1,l_2, ... $} \\
		&																	& \textbf{then} $L$ =  ''I want to see it at 5 pm'' & \\
		\bottomrule
	\end{tabular}
	\label{tbl:template}
	\vspace*{-3mm}
\end{table}

%\subsection{Generate Data from Template}

%\textbf{Template Definition}
\subsection{Template Definition}

Template $T$ is the input to our framework and is used to generate data. 
We define a template as: $T = (G, V, N)$,
%We define the input templates as $V = (\mathcal{T}_{rule}, \mathcal{T}_{nlg}, \mathcal{T}_{goal}, \mathcal{T}_{nlu})$. 
which includes user goals $G$, response strategies $V$ and natural language generation rules $N$.
An example of movie booking domain template is illustrated in Table\ref{tbl:template}.
%For example, considering a movie booking domain, the template of this domain is illustrated in Table\ref{tbl:template}. 

$G$ is a set of predefined user goals and defined as: 
$G=\{g_i\}_i^{\alpha}$,
where $\alpha$ is the number of goals. 
Each user goal $g$ is defined as $g=(C, R)$ 
which includes a set of user constraints $C$ and a set of user requests $R$ \cite{schatzmann2007agenda}. 

$V$ is a set of rules for response strategies, which is relatively easy to obtain \cite{Li2017}. 
It is defined as: 
$V=\{r_i\}_i^{\beta} =V_a \cup V_u $
, where $V_a$ and $V_u$ are response rules for user and agent respectively. 
For each $r \in V$, 
we define it as a function that maps dialogue context to response: 
\begin{equation}
\left\{
\begin{array}{lr}
r : f(A_1, A_2,  ... , A_{t-1}) \rightarrow A_t, & r \in V_a\\
r : f(A_1, A_2,  ... , A_{t-1}, G) \rightarrow A_t, &  r \in V_u\\
\end{array}
\right.
\notag
\end{equation}
where $A_i$ is the response for the $i_{th}$ turn. 
Specifically, as User and agent may take multiple actions in one turn, 
we define response $A$ as a set of single actions: $A = \{a_i\}_0^{k}$.
%
%Each rule $r$ yields actions on considering previous user and agent response $\langle A_1, A_2, A_3, ..., A_{t-1} \rangle$ (and user goal $G$). 
%User and agent may take multiple actions in one turn, 
%for example, user could request information and inform user constraint at the same time. 
%We define the turn $j$'s dialogue response action as a set of single actions $A_j = \{a_i\}_j^{K_j}$. 

$N=\{l_i\}_i^{\gamma} $ is a set of rules for natural language generation(NLG): 
%\[
%\begin{array}{c}
%N=\{l_i\}_i^{\gamma} \\
%l : f(a_1, a_2, a_3, ... , a_n, g) \rightarrow L \\
%\end{array}
%\]
Each rule $l$ maps actions and user goal to natural language $L$ and is defined as $l : f(a_1, a_2, a_3, ... , a_n, g) \rightarrow L$.

\begin{figure*}[t]
	\small
	\centering\includegraphics[width=1\columnwidth]{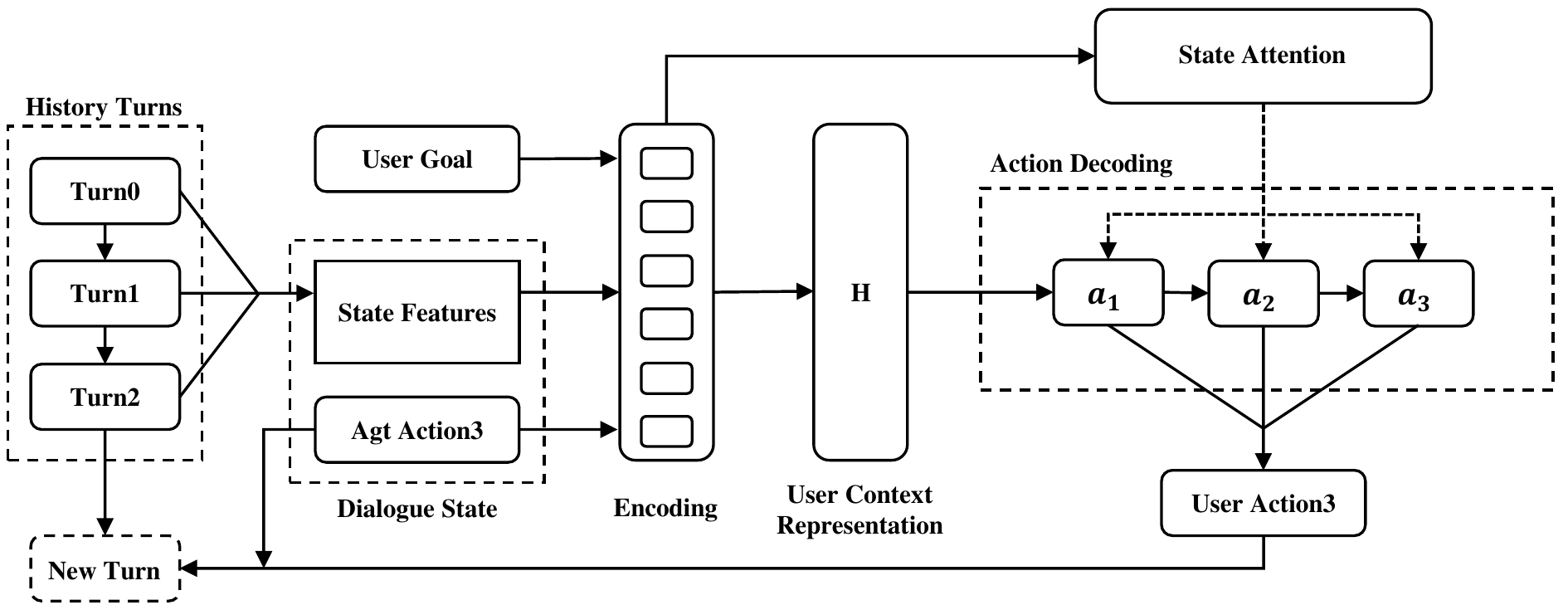}
	\caption{State2Seq user simulator.}\label{fig:state2seq}
\end{figure*}

\subsection{Data Generation}
The data generator generates conversation log with templates $T$ as input, as shown in Figure \ref{fig:framework}. 
There are two steps for the generation: 
\romannumeral1. Generate basic data with templates only. 
\romannumeral2. Generate diverse data with a built-in RL agent and templates. 

\textbf{Generate basic data}
To start from no corpus, 
we first collect some rule-based conversation as basic data, 
which is a common warm-up option \cite{Li2017}. 
When generating one dialogue, 
we first pick a user goal $g$ from $G$ as background and construct a random starting utterance. 
Then, for each turn, we search the suitable rule $r \in V$ to generate the response actions for user and agent. 
NLG rules $N$ is then used to render actions to utterance. 

\textbf{Generate diverse data}
On the basis of basic data, we generate diverse data. 
A built-in RL based agent $\mathcal{M}^*$ is used here. 
$\mathcal{M}^*$ is warmed up with basic data initially and further trained by interacting with $V_u$. 
When generating each dialogue, 
all process are same to basic data generation, 
except for agent response actions are given by $\mathcal{M}^*$'s policy. 

We enhance the data diversity by the following operations:
\romannumeral1. Leverage the RL exploration mechanism during data generation. 
\romannumeral2. Generate data with the built-in agents of different training stages. 
So both weak and strong policy are used, which allows collecting both clumsy and fluent dialogue.  

\begin{table}[t]
	\centering
	\renewcommand\arraystretch{1.5}
	\small
	\caption{Feature definition}
	\begin{tabular}{p{0.25 \columnwidth}p{0.7 \columnwidth}}
		\toprule
		\multicolumn{1}{c}{\textbf{Feature}} & \multicolumn{1}{c}{\textbf{Description}} \\ 
		\hline
		Constraint Status & Status about whether user constraint slots have been informed by user \\
%		\hline
		Request Status & Status about whether user request slots have been satisfied by agent \\
%		\hline
		Slot Consistency & Status of whether the slot values provided by agent are consistent to user constraints\\
%		\hline
		Dialog Status & Dialogue status of success, failed and no outcome yet\\
		\bottomrule
	\end{tabular}
	\label{tbl:feature}
	\vspace*{-3mm}
\end{table}

\subsection{User Model Training}
After collecting the dialogue data, 
we train the user simulator on it in supervised style. 
Given a dialogue context, 
the user model is trained to predict a set of user actions as response. 
Only user policy is learned from data here. 

\section{State2Seq User Simulator}
We aim to propose a user simulator that makes better use of dialogue context. 

User simulator mimics human responses to the dialogue system output. 
A user model predicts user response with dialogue context. 
Given user goal $g$ and dialogue history $<A_{t-1},  A_{t-2},  ... , A_{1}>$, 
it predict $t_{th}$ turn response as :
\[
A_t = \argmax_A{p(A \mid g, A_{t-1},  A_{t-2},  ... , A_{1})}
\]
where $A = \{a_i\}_0^{k}$. 
Following \cite{DBLP:conf/interspeech/AsriHS16}, we formulate the action selection as a sequence generation problem: 
%\[
%\begin{array}{l}
%p(A \mid g, A_{t-1},  A_{t-2},  ... , A_{1}) =  \\
%\quad\prod_{i}^n{p(a_i \mid a_1, ..., a_{i-1}, g, A_{t-1},  A_{t-2},  ... , A_{1})}
%\end{array}
%\]
\[
p(A \mid g, A_{t-1},  A_{t-2},  ... , A_{1}) =  \prod_{i}^n{p(a_i \mid a_1, ..., a_{i-1}, g, A_{t-1},  A_{t-2},  ... , A_{1})}
\]

However, dialogue history can be very long. 
So it is very hard for the user model to leverage history information directly. 
To remedy this, we extract the key information as dialogue state $S$ from the dialogue history. 
We then define the dialogue context as a combination of dialogue state and user goals. 

For better usage of dialogue context, 
we learn vector representations for each context items.
And attention mechanism is proposed to leverage context more clearly. 

Figure \ref{fig:state2seq} shows the structure of the State2Seq user simulator. 
The State2Seq model maintains a dialogue state $S$ for tracking the dialogue history. 
It uses an encoding module to provide vector representations for dialogue context items. 
For each turn, the model integrates those representations into a context representation $\textbf{H}$. 
Then the model decodes $\textbf{H}$ into a sequence of actions as output. 
Attention helps the model use dialogue context during decoding.

\subsection{Dialogue Context Representation}
%\textbf{Semantic Representation of Dialogue Context}

The main idea of dialogue context representation is to refine history information. 
Forgetting useless information is proven to be important for data-driven model. 

The dialogue context consists of dialogue state and user goal. 
Dialogue state $S$ includes last turn agent response and state features.
Following \cite{DBLP:conf/interspeech/AsriHS16}, we extract the state features explained in Table \ref{tbl:feature}. 
Constraint Status, Request Status, and Dialog Status are used to help the user simulator track the progress of the current conversation, 
and Slot Consistency allows the user simulator to correct the agent on wrong information.
Each feature is recorded as a status vector $\{status\}_0^m$, where $m$ is the number of slot types and  
$status$ could be 1, 0, -1 for active, irrelevant and negative.

%We propose an encoding module to learn semantic representation 
%for features, user goal and last turn agent response. 
%The representations of dialogue state and user goal are important for both the attention mechanism and 
%construction of user context representation $\mathbf{H}$. 

Attention mechanism relies on a good representation of items. 
Representation is also important for sequence decoding, as its need a good initial state $\mathbf{H}$. 
So we propose an encoding module to learn vector representations for context items.  
To help to learn of representation,
we share representations for the common slots in context items. 
Some negative status is rare in the corpus, which makes it hard to learn good representations. 
To remedy this, 
we only learn vector representations $\mathbf{e}$ for positive slot status, 
and then use the corresponding inverse vector $-\mathbf{e}$ to represent negative status.

H is obtained by dimension reduction of context representation, which can further forget irrelevant information.
Formally, given a agent response $A_{t-1}$, the user goal $g$ and current dialogue history $< A_{t-1},  A_{t-2},  ... , A_{1}>$, 
State2Seq updates dialogue state $S$ and represent context with vectors. $\mathbf{H}$ is then obtained as:
\[
\mathbf{H}=\mathbf{W}_c \cdot ([\mathbf{E}(A_{t-1}); \mathbf{E}(g); \mathbf{E}(S)]) + \mathbf{b}_c
\]

where $[;]$ denotes vectors concatenation and $\mathbf{E(x)}$ means fetching vector representations for items in $x$.

\subsection{Action Generation with Attention}
%\textbf{User Response Generation with State Attention}

%A real user for task-oriented dialogue would make multiple actions according to specific information of dialogue process. 
%To model the number variability of actions in one turn, 
%we formulate the user response as a sequence of actions following \cite{DBLP:conf/interspeech/AsriHS16}. 
%Neural user simulators are inherently diverse for multiple actions, such as seq2seq models \cite{DBLP:conf/interspeech/AsriHS16}. 
We formulate actions selection as sequence generation process with attention. 
Sequence generation provides diversity in selecting actions. 
Attention mechanism helps to use dialogue context information directly. 
Specifically, we generate responses by attending on items in user goal $g$, dialogue state $S$ and the last agent response $A_{t-1}$. 

During actions decoding, for each time step $t$,  LSTM provides hidden representation
$
\mathbf{h_k} = \text{LSTM}(\mathbf{h_{k-1}}, \mathbf{c_{k-1}}, \mathbf{x_k})
$
, where $\mathbf{h_k}$ denotes the LSTM hidden state at time step $k$, the $\mathbf{c_{k-1}}$ is cell state and $\mathbf{x}$ is input.

%We define the $\mathbf{Emb}$ as a set of embedding vectors for the user goal, the agent response and dialogue states:
%\[
%\mathbf{Emb} = E(g) \cup  E(S) \cup E(A_{t-1}) 
%\]
%where $E(X)$ means fetching the embedding vectors of all items in $X$. 
The attention weight of $i_{th}$ item in dialogue context is calculated as: 
$
\mathbf{att_i} = \frac{\text{exp}(\mathbf{e_i} \cdot \mathbf{h_k}) }{\sum_j \text{exp}(\mathbf{e_j} \cdot \mathbf{h_k})} 
$
where $\mathbf{e_i}$ is semantic representation vector of $i_{th}$ item.
Attention is used to calculate the decoding output 
%$\mathbf{h'_t}$:
$
\mathbf{h'_t} = \tanh(\mathbf{w_a} \cdot (\mathbf{att} \cdot \mathbf{Emb}) + \mathbf{b_a} \cdot \mathbf{h_k}) 
$
, where $\mathbf{Emb}$ is the representations of all items in context.

Then we model the distribution the time step $k$'s action $a_k$:
\[
P(a \mid g, S, A_{t-1}, a_{k-1}, ...,  a_{1}) = \text{Softmax}(\mathbf{W_p}\mathbf{h'_t} + \mathbf{b_p})  \\
\]

And the user action $a_k$ is predicted as:
\begin{equation*}
a_k = \argmax_a{p(a \mid g, S, A_{t-1}, a_{k-1}, ...,  a_{1})} \\ 
\end{equation*}

\section{Experiment}

\subsection{Dataset}

We used two datasets in our experiments: movie ticket booking data and restaurant reservation data.
%We mainly use the movie ticket booking dataset~\cite{Li2017}. 
The movie ticket booking data is an open task-completion dialogue dataset proposed by \cite{Li2017}. 
For each dialogue, the system gathers information about the customer’s desires and books the movie tickets. 
The success or failure of dialogue is assessed based on (1) whether a movie is booked, and (2) whether the movie satisfies the user's constraints. 
The data includes 11 dialogue acts, 29 slots, 277 user goal templates. 
Rule templates for response strategy and NLG are also included in \cite{Li2017}'s work.

To test the method's generation ability,  
we also build a dataset for restaurant reservation domain. 
In each dialogue, the user reserves a table under his/her requirements.
The data includes 11 dialogue acts, 24 slots and 184 user goal templates. 
We design rule templates for response strategy and NLG based on the ones in \cite{Li2017}'s work. 

\subsection{Evaluation}
%Evaluation of user simulators is a problem far from solved \cite{SCHATZMANN2006}.
We evaluate a user simulator by: 
\romannumeral1. evaluation of the user simulator itself. 
\romannumeral2. evaluation of the policy trained with it. 
%Two Agents are leveraged as an evaluation tool to perform the conversation test. 

\textbf{Evaluation of Agent Policy}
The main value of simulator is to train agent policy. 
We use both human and automatic evaluation for policy here.
A DQN model is used as the agent to learn the policy. 

We adopt cross-model evaluation proposed by \cite{JostSchatzmannMatthewN.Stuttle2005}. 
$\mathcal{N}$ user models are first used to train $\mathcal{N}$ policies. 
Each policy is then tested against $\mathcal{N}$ different user simulators. 
Finally, we calculate the average of $\mathcal{N} \times \mathcal{N}$ scores. 
A policy trained by a good simulator can still perform well on poor simulators \cite{JostSchatzmannMatthewN.Stuttle2005,DBLP:conf/sigdial/KreyssigCBG18}. 
A higher average score indicates a better simulator ability for training agent. 
For metric, we use success rate, average reward and average turn number, 
which have been widely accepted as a standard metric of multi-turn agent 
\cite{DBLP:conf/acl/GaoWPLL18,DBLP:journals/corr/abs-1711-11023,Li2017}. 

\textbf{Evaluation of User Simulator}
We evaluate the user simulator itself from two aspects.

Firstly, following \cite{DBLP:conf/interspeech/AsriHS16}, 
we evaluate the accuracy of predicting actions. 
F-score is used as the metric. 

Secondly, a user simulator's generalization ability is also important. 
It has more tolerance for exploration of training agent. 
%It can be evaluated through conversations with agent. 
We compare different user simulators' such ability by making conversation against a same rule agent.
User model with better ability should achieve more success rate. 

\vspace*{-3mm}
\subsection{Model Details}
For the State2Seq model, 
we set embedding size as 300.  
%and share the embedding layer among user goal, agent response and dialogue state. 
We use a 2 layer LSTMs for decoding with hidden state size of 256.
During the training, 
we set the batch size as 32, a dropout as 0.8 
and teacher forcing rate of 0.5. 
The learning rate is set as 0.001 and we set a learning rate decay of 0.9. 
%b_s32-h_d256-m_e30-d_o0.8-dep3-lr0.001-lr_d0.9-t_f0.5

For the RL agent model used in data generation and evaluation, 
we use the DQN model. 
We set experience pool size as 1000, hidden layer size as 80. 
%and DQN's $\gamma$ as 0.9.  
Experience replay redesigned for dialogue setting is applied. 
We use $\epsilon$-greedy exploration of 0.01. 
The learning batch size is 16. 
We use 100 warm-up epochs and 500 training epochs. 
Model simulates 100 dialogues for each epoch. 

We also simply extend our model by replacing the sequence decoder with a multi-label classifier. 
We name it as State2MLC. State2MLC takes dialogue state, user goal, agent response as input and predict multiple actions. 
State2MLC is trained with Multi-Label Soft Margin Loss.

\vspace*{-3mm}
\subsection{Baselines}
We compared with the following baselines:

\begin{itemize}
	\item \textbf{Seq2Seq} is user simulator proposed by \cite{DBLP:conf/interspeech/AsriHS16}. 
	It extracts history turns' features as input sequence and decodes action sequence. 
	
	\item \textbf{Seq2Seq-att} is based on Seq2Seq model and 
	adds attention mechanism over input sequence. 
	
	\item \textbf{Agenda} based user simulator is proposed by \cite{schatzmann2007agenda}.  
	It is corpus free and generates user response by maintaining a user agenda with rule. 
	We use the agenda user simulator provided by \cite{Li2017}. 
	%	Here, the agenda based user simulator could also be viewed as the template based user simulator. 
	
	%\[
	%loss(x,y)=−\sum_iy[i]∗log((1+exp(−x[i]))−1)+(1−y[i])∗log(exp(−x[i])(1+exp(−x[i])))
	%\]
\end{itemize}

\begin{table*}[t]
	\centering
	\footnotesize
	\renewcommand\arraystretch{1.2}
	\caption{Evaluation of agent policy trained by different user model. Results above dash-line are from our model, which achieve best performance in most task.}
	\begin{tabular}{lcccccc}
		\toprule
		\multirow{2}{*}{Model} & \multicolumn{3}{c}{Movie Booking} & \multicolumn{3}{c}{Restaurant Reservation}  \\
%		\cline{2-7}
		\cmidrule(lr){2-4}
		\cmidrule(lr){5-7}
		& Avg. Succ. & Avg. Rwd.& Avg. Turns & Avg. Succ. & Avg. Rwd.& Avg. Turns\\
		\hline
		State2MLC  & 0.487 & 8.55 & \textbf{21.82} & 0.305 & -17.22 & 29.69 \\
		State2Seq & \textbf{0.551} & \textbf{14.17} & 25.85 & \textbf{0.524} & \textbf{11.77} & 24.21 \\
		\hdashline
		Seq2Seq & 0.412 & -3.49 & 27.77 & 0.501 & 6.23 & 29.83 \\
		%		Seq2Seq \cite{DBLP:conf/interspeech/AsriHS16} & 0.4116 & -3.492 & 27.77 \\
		Seq2Seq-Att & 0.430 & -2.59 & 30.39 & 0.514 & 9.67 & 26.20 \\
		Agenda  & 0.438 & -2.88 & 32.88 & 0.508 & 10.88 & \textbf{22.17} \\   % old version
		%		Agenda \cite{li2016user} & 0.438 & -2.880 & 32.88 \\   % old version
		\bottomrule
	\end{tabular}
	\label{tbl:cross_eval}
\end{table*}

\begin{table}[t]
	\centering
	\renewcommand\arraystretch{1.2}
	\footnotesize
	\caption{Human evaluation of trained agent}
	\begin{tabular}{lccc}
		\toprule
		Model & Avg. Succ. & Avg. Rwd.& Avg. Turns \\
		\hline
		State2Seq (Ours) & \textbf{0.778} & \textbf{53.88} & \textbf{11.88} \\
		Agenda & 0.571 & 22.29 & 14.57 \\
		\bottomrule
	\end{tabular}
	\label{tbl:human_eval}
	\vspace*{-5mm}
\end{table}

\begin{table}[t]
	\centering
	\renewcommand\arraystretch{1.2}
	\caption{
		Analysis of agent policy performance during training on movie domain. 
		At beginning of each policy training epoch, 
		we test the policy's performance. 
		The results reflect the models' overfitting to training set. 
	}
	\begin{tabular}{l|ccc}
		\toprule
		Model & Succ. & Avg. Rwd.& Avg. Turns \\
		\hline
		State2MLC (Ours)  & 0.628 & 26.19 & 16.36 \\
		State2Seq (Ours) & 0.800 & 48.39 & 17.23 \\
		\hdashline
		Seq2Seq & 0.462 & 2.98 & 28.92 \\
		Seq2Seq-Att & 0.480 & 2.11 & 32.98 \\
		Agenda  & \textbf{0.814} & \textbf{50.36} & \textbf{15.66} \\
		\bottomrule
	\end{tabular}
	\label{tbl:AgentRes}
\end{table}

\begin{table*}[t]
	\centering
	\renewcommand\arraystretch{1.2}
	\footnotesize
	\caption{Evaluation of user simulator model. }
	\begin{tabular}{lcccccc}
		\toprule
		\multirow{3}{*}{Model}& \multicolumn{2}{c}{Action Accuracy} & \multicolumn{4}{c}{Generalization Ability Test} \\
%		\cline{2-7}
		\cmidrule(lr){2-3}
		\cmidrule(lr){4-7}
		& Movie & Restaurant & \multicolumn{2}{c}{Movie} & \multicolumn{2}{c}{Restaurant} \\
		\cmidrule(lr){2-2}
		\cmidrule(lr){3-3}
		\cmidrule(lr){4-5}
		\cmidrule(lr){6-7}
		& F1 & F1 & Avg. Succ. & Avg. Rwd. & Avg. Succ. & Avg. Rwd. \\
		\hline
		State2MLC (Ours) & 0.704 &  \textbf{0.695} & \textbf{0.442} & \textbf{6.06} & 0.436 & 5.34 \\
		State2Seq (Ours) & \textbf{0.711} & 0.683& 0.400 & 5.51 & \textbf{0.484} & \textbf{11.09} \\
		\hdashline
		Seq2Seq  & 0.692 & 0.662 & 0.063 & -31.87 & 0.126 & -31.87 \\
		Seq2Seq-Att  & 0.705 & 0.677 & 0.000 & -46.99 & 0.000 & -46.99 \\
		Agenda & N/A & N/A & 0.392 & 0.04  & 0.410 & 2.20 \\
		\bottomrule
	\end{tabular}
	\label{tbl:UserRes}
\vspace*{-5mm}
\end{table*}

\vspace*{-5mm}
\subsection{Performance of Agent Policy}
We compare the policy trained by different user simulators with cross-model evaluation. 

\textbf{Results on movie booking data}
Table \ref{tbl:cross_eval} shows the evaluation results on movie booking domain. 
The results show that the policy obtained by our model outperforms baselines. 
%on all metrics. 

On average success rate, policy trained by our State2Seq model outperforms the Agenda model for $6.36$ points. 
As the Agenda \cite{Li2017} model is rule-based, 
%this shows that our framework builds better user simulator than rules. 
the main difference between State2Seq and it is that State2Seq has more diverse responses. 
This demonstrates that user simulator diversity improves policy ability for finishing task and generalization. 
Policy trained by our model outperforms the Agenda model on success rate and average reward. 
%But it lags a bit on average number of turns compared to Agenda model. 
%This is because our policy shows more patient to finish task. 
%It is observed that the proposed user simulator is always more patient to a poor agent response, 
%which would increase the length of dialogue rather than closing dialogue. 

The results show that the policy trained by other statistical methods underperform the Agenda based model. 
Because those user simulators lack for response accuracy, 
which will mislead and confuse the policy training. 
Comparing Seq2Seq-att to Seq2Seq, the results show the effectiveness of attention mechanism. 
And State2Seq's improvements over the Seq2Seq-att show that 
the refined context representation of the State2Seq model does help the response generation. 

\textbf{Results on restaurant data}
Table \ref{tbl:cross_eval} shows the cross evaluation results for restaurant reservation domain.
The results show our method could work consistently well in different domains. 
Most models score higher on data in the restaurant domain, 
because the field is relatively simpler and has fewer slots. 
The State2MLC model does not perform well. 
This is due to the fact that State2MLC model has a much simpler structure, 
so it is likely to overfit to generated data in a simple domain and limits the generalization ability of the policy. 

\textbf{Human evaluation}
We perform a human evaluation on the movie domain, 
and each agent is tested by chatting with 2 domain experts for 50 dialogues. 
Table \ref{tbl:human_eval} shows that 
the agents trained in our model can be better adapted to the real situation. 

\textbf{Analysis on policy training process}
Table \ref{tbl:AgentRes} shows the evaluation of training process. 
%Before each training epoch, 
%we perform testing with user goals sampled from the test goal set and report the averaged scores of all epochs.  
We perform testing at each training epoch, and report the averaged score. 
The policy is evaluated against the environment for training. 
Policy trained by our model outperforms the ones from statistical user simulator, 
which reflects that our user simulator improves the training performance of agent policy. 
It is because our user simulator has better generalization to respond to agent's unreasonable actions in the early training stage. 
On the other hand, diversity helps RL algorithm training. 
Agenda achieve good scores as rule environment is relatively easy for overfitting.

\vspace*{-3mm}
\subsection{Performance of User Simulator}
User model's performance is mainly reflected by the ability of predicting user responses. 

\textbf{Action accuracy.}
Table \ref{tbl:UserRes} shows the models' accuracy of predicting user actions. 
Our model achieves the best performance, outperforming the seq2seq model \cite{DBLP:conf/interspeech/AsriHS16} for $1.9$ points on F-score. 
%These results show that our model has a better ability of generating reasonable actions. 
The results also show a correlation between the evaluation of action prediction and the agent policy's performance, 
%which demonstrates that a good user simulator is useful for training agents. 
which demonstrates that user simulator quality affects agents performance. 

The improvements mainly come from two aspects: 
Firstly, the attention mechanism provides specific context information.
Secondly, refined context representation filters the useless information. 
The Seq2Seq-att model outperforms the Seq2Seq model by adding the attention mechanism. 
This reflects the effectiveness of the attention mechanism. 
By comparing State2Seq to Seq2Seq-att, 
improvement shows that due to forgetting mechanism, 
refined context representation is more effective than sequence encoder. 
%For context representation, a sequence encoder tries to capture all dialogue information of each turn equally, 
%and each turn contains heavy information of dialogue state and user goal state.  
%But State2Seq represent dialogue context with last turn agent response, dialogue state feature and user goal, 
%which provides sufficient but summarized information. 

\textbf{Analysis of generalization ability}
The generalization ability is also important for a user simulator in agent training. 
We compare different user simulators' generalization by making them chat with a same rule-based agent. 
Table \ref{tbl:UserRes} shows the results and 
State2Seq and MLC2Seq are optimal in all user simulators. 
Our best model outperforms the Agenda for $5.0$ and $7.4$ points success rate on the 2 domains. 
%As the user simulators are trained on data generated from the template and Agenda equals to the strategy rules in template, 
As the user simulators are trained to mimic user rules in the template, 
the improvement reflects that our framework can generate new diverse dialogues data to avoid user simulator overfitting to user response strategy in the template.
Other user simulators perform worse on this test, 
we address this to the fact that these methods are less accurate in generating user actions
and rule-based agent has a low allowance for response error. 

\begin{table}[t]
	\centering
	\scriptsize
	\renewcommand\arraystretch{1.2}
	\caption{
		Case study of the difference between our user simulator 
		%		\wx{It's longer? And the first two turns are duplicate?} 
		and agenda user simulator. 
		Here, user is requesting for ticket and  theater, and user's constraints are \{ movie name: deadpool, city: Seattle ,  num: 2, date: tomorrow \}
	}
%	\begin{tabular}{p{0.2\columnwidth}p{0.75 \columnwidth}}
%		\toprule
%		\multicolumn{2}{c}{\textbf{User Goal}} \\
%		\hline
%		\textbf{Request: } & ticket, theater \\
%		\textbf{Constraint:} & moviename: deadpool, city:Seattle ,  num: 2, date: tomorrow \\
%		\hline
%	\end{tabular}
%	\begin{tabular}{p{0.25\columnwidth}p{0.7 \columnwidth}}
%		\toprule
%		 \multicolumn{1}{c}{\textbf{User Request}} & \multicolumn{1}{c}{\textbf{User Constraint}} \\
%		\cmidrule(lr){1-1}
%		\cmidrule(lr){2-2}
%		ticket, theater & moviename: deadpool, city:Seattle ,  num: 2, date: tomorrow \\
%		\hline
%	\end{tabular}
	\begin{tabular}{p{0.95 \columnwidth}}
		\toprule
		\multicolumn{1}{c}{\textbf{Agenda User Simulator}} \\
		\hline
		\tabincell{p{0.4 \columnwidth}p{0.55 \columnwidth}}{
			...  \\
			usr: Which theater is available? & act:req, req slots: \{theater\} \\
			agt: Which theater would you like? & act:req, req slots: \{theater\} \\
			usr: Which theater is available? & act:req, req slots: \{theater\} \\
			agt: Which theater would you like? & act:req, req slots: \{theater\}\\
			(loop...) \\
		} 
		\\
		\hline
		\multicolumn{1}{c}{\textbf{Our User Simulator}} \\
		\hline
		\tabincell{p{0.4 \columnwidth}p{0.55 \columnwidth}}{
			... \\
			usr: Which theater is available? & act:req, req slots:\{theater\} \\
			agt: Which theater would you like? & act:req, req slots:\{theater\} \\
			usr: I want to watch at Seattle. & act:inform, inform slots:\{city: Seattle\}\\
			agt: Seattle is available. & act:inform, inform: \{city: Seattle\}\\
			usr: Which theater is available? & act:req, req slots: \{theater\}\\
			agt: The Pearl Theater is available. & act:inform, inform slots: \{theater: The Pearl Theater\} \\
			... \\
		}
		\\	
		\bottomrule
	\end{tabular}
	\label{tbl:case-study}
	\vspace*{-5mm}
\end{table}

\begin{figure}[t]
	\scriptsize
	\centering
	%	\tabincell{p{0.43\columnwidth}p{0.55\columnwidth}}{
	%%		\hline
	%%		\multicolumn{2}{c}{\textbf{Dialogue Context} }\\
	%%%		\hdashline
	%%		\hline
	%		agt: How many tickets do you need? & act:req, inform S:\{\},req S:\{\# of people\} \\
	%		usr: I want 2 tickets please!  & act:inform, inform S:\{\# of people:2\}, req S:\{\}\\
	%%		\hdashline
	%	}
	\begin{tikzpicture}
	\draw (10,0 ) node[inner sep=0] {\includegraphics[width=0.7\columnwidth, trim={0.4cm 0.4cm 0.4cm 0.3cm}, clip]{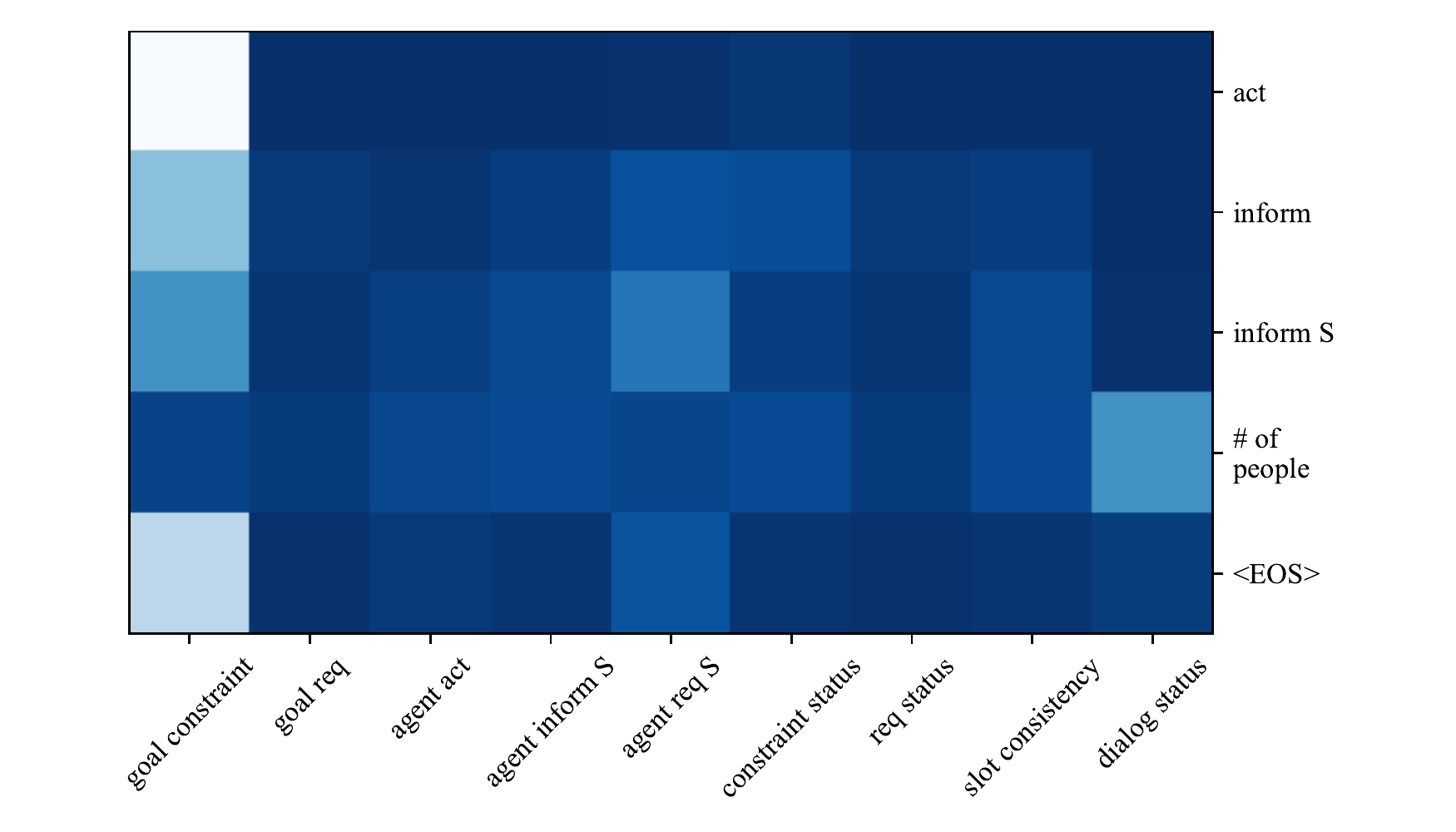}};
	\end{tikzpicture}
	\tabincell{p{0.41\columnwidth}p{0.55\columnwidth}}{
		%		\hline
		%		\multicolumn{2}{c}{\textbf{Dialogue Context} }\\
		%%		\hdashline
		%		\hline
		Context: & Actions: \\
		agt:How many tickets do you need? & act:req, inform S:\{\}, req S:\{\# of people\} \\
		usr:I want 2 tickets please!  & act:inform, inform S:\{\# of people:2\}, req S:\{\}\\
		%		\hdashline
	}
	\caption{Visualization of attention. 
		The vertical axis is the generated user response and 
		the horizontal axis is the dialogue context. The lighter color means higher attention. 
		S denotes slots. 
	}
	\label{fig:attention}
	\vspace*{-5mm}
\end{figure} 

%\subsection{Case Study}
\vspace*{-2mm}
\subsection{Case Study and Visualization}
To find out the difference between rule-based model and the proposed model. 
We perform case study on dialogues between the user simulator and an agent. % trained by it. 
The comparison is shown in Table \ref{tbl:case-study}.
When agents and users are unable to satisfy each other, 
the fixed rules of Agenda user simulator may be trapped in the loop shown in the example. 
But the response from our user simulator is diverse and uncertain. 
It can try other actions to jump out of the endless loop. 
The difference shows that our proposed framework for building user simulators has successfully improved the response diversity.

To demonstrate the effectiveness of the attention mechanism over context, 
we provide a visualization example shown in Figure \ref{fig:attention}.
%The blue-level of a cell indicates the attention scores. 
The figure shows that the attention mechanism successfully learns the 
correlation between the generated actions and the context.
Specifically, when generating the dialogue action of \texttt{inform} and the inform slots, 
the proposed model pays a higher attention to context items of \texttt{sys\_request\_slots} and \texttt{user\_goal}. 

\vspace*{-2mm}
\section{Related Work}
There is little work solving the data insufficient problem of user simulator.
\cite{DBLP:conf/sigdial/KreyssigCBG18} proposed a user simulator that generates user utterance directly, 
which could ease the effort of user semantic annotation. 
However, their work is corpus-based and still needs a large corpus to train on it. 
Other data scarcity problems for task-oriented dialogue system are also investigated.
\cite{DBLP:journals/corr/abs-1805-04803} leverage the idea of zero shot learning \cite{NIPS2009_3650}.
They solve the problem of dialogue generation for a new domain by mapping actions to latent space.  
\cite{DBLP:conf/interspeech/KurataXZ16} and \cite{hou2018sequence} 
provided data augmentation methods for language understanding. 
\cite{shah2018building} proposed to build new domain agent efficiently by machine self-play and crowdsourcing.

The first user simulator was proposed by \cite{eckert1997user}. 
%Traditional user simulators are built with handcraft rules \cite{schatzmann2007agenda,Li2017},  
%Data driven user simulators learn user behaviors in datasets with statistical models. 
%\cite{cuayahuitl2005human,georgila2005learning,pietquin2006probabilistic,DBLP:conf/interspeech/AsriHS16}. 
There are two kinds of user simulators in terms of working levels. 
User simulators of semantic level interact to agent with dialogue acts and corresponding slot-value pairs.  \cite{schatzmann2007statistical,chandramohan2011user,DBLP:conf/interspeech/AsriHS16,DBLP:conf/acl/GaoWPLL18}. 
User simulators of utterance level communicates to agent with utterance directly \cite{jung2009data,DBLP:conf/sigdial/KreyssigCBG18,liu2017iterative}. 
Our user simulator can work on both of the two levels.

%The user simulator closest to our model is a seq2seq user simulator proposed by \cite{DBLP:conf/interspeech/AsriHS16}, 
%which extracts the feature of each turn as input for encoding and directly decodes user actions as user response. 
%\cite{DBLP:conf/interspeech/AsriHS16} proposes an RNN based method that allows for modeling dialogue history and a variable number of actions. 
%%\cite{DBLP:conf/sigdial/KreyssigCBG18} proposed a seq2seq user simulator that generates user utterance directly to get rid of annotation of user action. 
%However, these models can not concentrate on specific dialogue information and 
%simple RNN model is a poor surrogate for long dialogue history representation \cite{DBLP:journals/corr/BahdanauCB14}. 
%In contrast, our model leverages attention mechanism to make better representation for dialogue information and help response generation. 

\vspace*{-2mm}
\section{Conclusion}
In this paper, 
we study the problem of building user simulators for 
task-oriented dialogue from templates with no corpus. 
%Traditional user simulator lacks for response diversity and data-driven user simulators provide diversity but data scarcity. 
%Our method combines the above two ideas of building user simulator.
We solve the data scarcity and increase simulator response diversity by proposing a corpus-free framework.
In our framework, we generate diverse data with only templates and trains a data-driven user simulator on it. 
To predict user response more accurately, 
we proposed a novel State2Seq user model. 
It predicts user response with attention on refined dialogue context representations. 
%The model also learn a refined context representations to do better prediction. 
Experiment results show that with more response diversity, our user simulator improves the agent policy by 6.36\% success rate. 
Attention and refined context representation help the State2Seq model outperform Seq2Seq baseline for 1.9 F-score.
%State2Seq user model performs better than all other baselines. 
\vspace*{-2mm}
\section*{Acknowledgments}
\vspace*{-2mm}
We are grateful for helpful comments and suggestions from the anonymous reviewers.  
This work was supported by the National Natural Science Foundation of China
(NSFC) via grant 61632011, 61772153 and 61772156.

 \bibliographystyle{splncs04}
 \bibliography{submission}
\end{document}